\documentclass[11pt]{article}

\usepackage[utf8]{inputenc}
\usepackage[margin=1in]{geometry}
\usepackage{amsmath,amssymb}
\usepackage{graphicx}
\usepackage{booktabs}
\usepackage{multirow}
\usepackage{hyperref}
\usepackage{xcolor}
\usepackage{enumitem}
\usepackage{microtype}
\usepackage{subcaption}
\usepackage{float}
\usepackage[numbers,super,sort&compress]{natbib}
\usepackage{setspace}

\DeclareCaptionLabelFormat{ncs}{\textbf{#1~#2~|}}
\DeclareCaptionLabelFormat{extdata}{\textbf{Extended Data Fig.~#2~|}}
\DeclareCaptionLabelFormat{extdatatab}{\textbf{Extended Data Table~#2~|}}
\newcommand{\extdatafigcaptions}{%
  \captionsetup[figure]{labelformat=extdata, labelsep=space, name={}}%
  \captionsetup[table]{labelformat=extdatatab, labelsep=space, name={}}%
}

\DeclareCaptionLabelFormat{suppfig}{\textbf{Supplementary Fig.~#2~|}}
\DeclareCaptionLabelFormat{supptab}{\textbf{Supplementary Table~#2~|}}
\newcommand{\suppfigcaptions}{%
  \captionsetup[figure]{labelformat=suppfig, labelsep=space, name={}}%
  \captionsetup[table]{labelformat=supptab, labelsep=space, name={}}%
}
\hypersetup{colorlinks=true,linkcolor=blue!70!black,citecolor=blue!70!black,urlcolor=blue!70!black}
\graphicspath{{figures/}}

\newcommand{\RnaOnlyOmicsR}{0.651}
\newcommand{\LincsGlobalRDelta}{-0.058}
\newcommand{\LincsPerDrugDelta}{+0.001}
\newcommand{\CellMeanR}{0.645}
\newcommand{\KFiftyR}{0.701}
\newcommand{\KFiftyLift}{+0.056}
\newcommand{\ErkMapkAllDrug}{0.427}
\newcommand{\ErkMapkMoaR}{0.723}
\newcommand{\ErkMapkProfileConcordance}{0.565}
\newcommand{\ErkMapkProfileConcordanceStd}{0.231}
\newcommand{\EgfrAllDrug}{0.425}
\newcommand{\EgfrMoaR}{0.799}
\newcommand{\DrugMeanOracleR}{0.837}
\newcommand{\PasoInflation}{+0.042}
\newcommand{\PasoBestFold}{0.751}
\newcommand{\MeasurementCeiling}{0.754}
\newcommand{\MoaOnehotGlobalRDelta}{+0.041}
\newcommand{\CtrpvTwoEgfrMoaDelta}{+0.371}
\newcommand{\CtrpvTwoKShotLift}{+0.051}
\newcommand{\BeatAmlKZeroR}{0.453}
\newcommand{\BeatAmlKFiftyR}{0.521}

\title{\Large\bfseries Training distribution determines the ceiling\\
of drug-blind cancer sensitivity prediction}

\author{Taekyung Heo\\[0.3em]
\small\texttt{taekyung.cs@gmail.com}}

\date{}

\begin{document}
\doublespacing
\maketitle

\section*{Abstract}

\noindent Precision oncology requires predicting which drugs will suppress a specific tumor from its molecular profile, but drug-blind sensitivity prediction has plateaued despite increasingly complex drug representations. Here we show that this stagnation reflects a metric artifact rather than a representational bottleneck. The standard benchmark, global Pearson $r$, is dominated by between-drug potency differences that a trivial drug-mean predictor captures without any cell-specific learning. Per-drug Pearson $r$, which isolates within-drug cell ranking, reveals that no drug encoding improves over cell-only features across four independent datasets. A controlled experiment channeling mechanism-of-action identity as either a drug feature or a training-distribution constraint identifies the cause. Supplying MoA as a feature yields negligible benefit, whereas using it to stratify training raises per-drug $r$ substantially for targeted kinase inhibitors, because pan-cancer co-training suppresses pathway-specific sensitivity signals. Mechanism-stratified training and response matching from pilot observations provide two deployable strategies that together recover the principal sources of predictive gain in drug-blind sensitivity prediction.

\section*{Introduction}

Predicting which drugs will suppress a specific tumor from its molecular profile is the central computational challenge of precision oncology\cite{costello2014community,adam2020machine}. In the drug-blind setting, where models must generalize to compounds absent from training, a decade of increasingly sophisticated drug encoders has converged on a ceiling of global Pearson $r \approx 0.45$--$0.55$\cite{zhu2022tgsa,han2026dispa}, raising the question of whether drug representations are fundamentally limited or merely underspecified.

Recent evidence reframes this debate. A drug-mean oracle achieves global Pearson $r = 0.85$ on GDSC\cite{specgame2025}, and removing the drug branch from deep models does not degrade performance in cancer-blind evaluation\cite{branson2025understanding}. We show that the same null extends to the drug-blind setting under per-drug $r$ (Fig.~\ref{fig:decomp}b). These observations indicate that global $r$ conflates two signals: between-drug potency ranking and within-drug cell ranking. Per-drug Pearson $r$, computed within each test drug across cell lines, isolates cell ranking and more directly addresses the clinical question of which patients respond to a given compound.

Under per-drug $r$, no drug representation produces more than a negligible improvement over cell-only features. This holds for structural encodings (Morgan fingerprints\cite{rogers2010extended}, ChemBERTa\cite{chithrananda2020chemberta}), functional encodings (LINCS L1000 perturbation signatures\cite{subramanian2017next}), and drug-target vectors, and is confirmed independently in drug-blind evaluation\cite{herbert2026monotherapy,bernett2025dreval} and in the drug-seen regime\cite{carli2025cellhit}. We replicate the null in both ridge regression and an 11.7M-parameter Transformer encoder. Reported ceilings above this null are explained by test-set checkpoint selection in several prominent models, consistent with widespread evaluation leakage documented across the field\cite{asiaee2026widespread}.

\begin{figure}[!b]
    \centering
    \includegraphics[width=\textwidth]{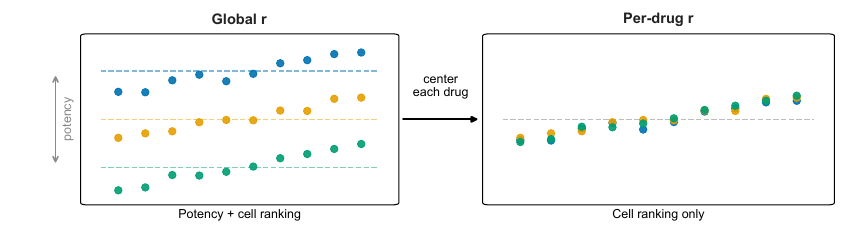}
    \caption{\textbf{Global $r$ conflates two independent signals. Per-drug $r$ isolates within-drug cell ranking.}
    Three drugs (colored) span different mean sensitivity levels (potency) while sharing the same relative cell-line ordering. Global Pearson $r$ (left) reflects both potency differences and cell ranking. Removing each drug's mean (per-drug $z$-scoring) yields per-drug $r$ (right), which isolates cell ranking.}
    \label{fig:framework}
    \end{figure}

Here, we address whether this failure is informational (the model lacks the right drug encoding) or distributional (the model trains on the wrong drugs). A controlled experiment channels the same mechanism of action (MoA) information through two routes: as a drug feature or as a training distribution (Fig.~\ref{fig:moa}). Training distribution alignment dominates representation across both ridge regression and an 11.7M-parameter Transformer encoder, and two practical strategies exploit this to improve within-drug cell ranking across independent datasets.
\section*{Results}

\subsection*{Per-drug evaluation reveals a hidden ceiling}

The standard drug-blind metric, global Pearson $r$ over all test pairs, conflates two signals: between-drug potency ranking and within-drug cell sensitivity ranking\cite{specgame2025}. A drug-mean oracle that predicts each compound's average IC$_{50}$ regardless of cell line already surpasses the reported drug-blind ceiling by a wide margin (global $r = \DrugMeanOracleR$ vs.\ the published ceiling of ${\sim}0.51$), demonstrating that the ceiling reflects failure to predict drug potency for unseen compounds, not failure to rank cells.

\begin{figure}[!htbp]
\centering
\includegraphics[width=\textwidth]{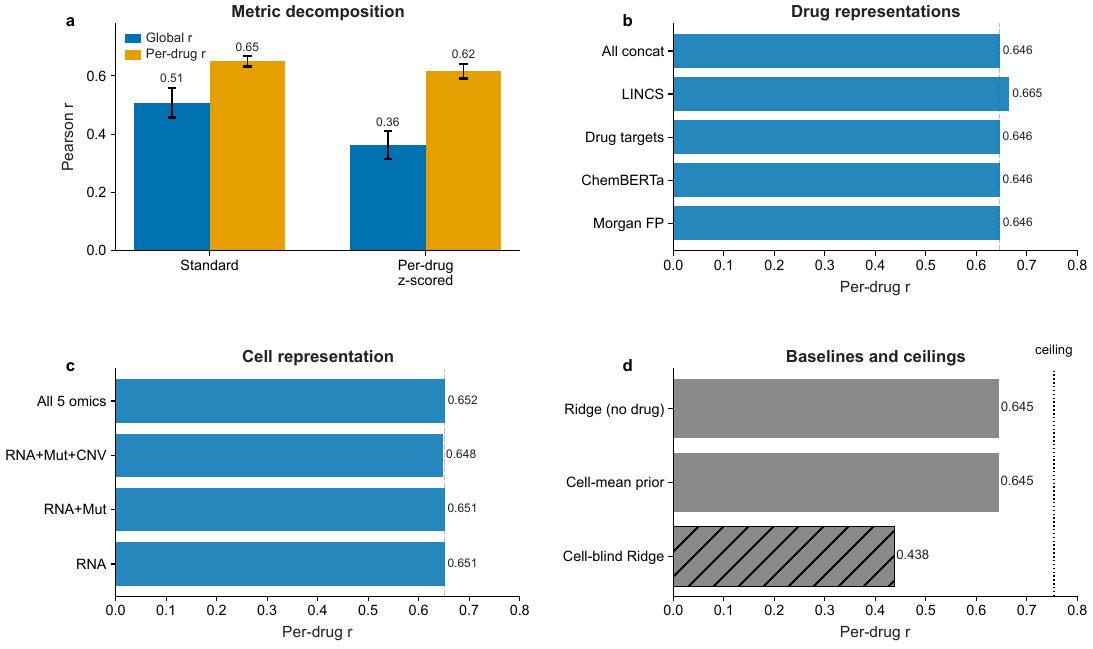}
\caption{\textbf{Per-drug $r$ reveals a ceiling invisible to global $r$, and no drug representation improves within-drug cell ranking.}
\textbf{a}, Global vs.\ per-drug $r$ under standard training and $z$-scoring.
\textbf{b}, Per-drug $r$ across drug feature types (structural and functional), all within $\Delta \leq +0.001$ of the no-drug-features baseline (dashed). LINCS is evaluated on its 104-drug covered subset (within-subset $\Delta = +0.001$).
\textbf{c}, Omics ablation. RNA-seq alone achieves per-drug $r = \RnaOnlyOmicsR$; mutations and further modalities each add $\leq +0.001$.
\textbf{d}, Per-drug $r$ for three reference predictors: cell-blind ridge (out-of-distribution cells), cell-mean prior ($K=0$ zero-shot), and ridge without drug features (drug-blind standard). Dotted line: within-assay replicate ceiling ($r = \MeasurementCeiling$).}
\label{fig:decomp}
\end{figure}

Global $r$ decomposes into a weighted sum of between-drug and within-drug components (Supplementary Note~1). In GDSC2, between-drug variance dominates because drug means span an order of magnitude more than within-drug variation, so global $r$ is driven almost entirely by scale prediction (Fig.~\ref{fig:decomp}a).

Per-drug Pearson $r$, the correlation within each test drug across cell lines averaged over drugs, isolates cell ranking. Training with per-drug $z$-scored targets collapsed global $r$ while preserving per-drug $r$, confirming that the two signals are independent (Fig.~\ref{fig:decomp}a). Cell-blind CV on held-out cell lines inverts the conventional ranking: familiar cells enable accurate ranking without drug information, but unfamiliar cells cannot be ranked even when the drug is known (Fig.~\ref{fig:decomp}d).

Cross-assay replicate concordance between GDSC1 and GDSC2 provides an empirical ceiling of $r = \MeasurementCeiling$ (9 anchor drugs, per-drug range 0.57--0.93). The zero-shot baseline already reaches 86\% of this ceiling (Fig.~\ref{fig:decomp}d).

\subsection*{Evaluation artifacts in three prominent models}

PASO\cite{wu2025paso} reported drug-blind $r = 0.745$. Examination of the publicly available implementation revealed checkpoint selection by maximizing \emph{test-set} Pearson $r$ each epoch with no validation holdout. DeepCDR\cite{liu2020deepcdr} and DrugCell\cite{kuenzi2020predicting} showed the same pattern. Under a fair protocol with a validation holdout, test-set leakage and best-fold selection together account for the gap between the fair mean and the published figure (Extended Data Fig.~1). Leakage was confirmed in 23 of 32 audited methods across the field\cite{asiaee2026widespread}. We therefore ask, under a corrected evaluation protocol, whether any drug encoding can improve per-drug $r$.

\subsection*{Drug representations do not improve within-drug ranking}

We evaluated all major drug representation types under ridge regression with 10-fold drug-blind CV on PASO splits. No encoding improved per-drug $r$ beyond $\Delta = +0.001$ of the cell-only baseline (Fig.~\ref{fig:decomp}b, Extended Data Table~1). This null held for Morgan fingerprints, ChemBERTa embeddings, drug-target vectors, and LINCS L1000 perturbation signatures\cite{subramanian2017next}. The LINCS result is particularly informative because these signatures encode a drug's transcriptional effect rather than its chemical structure, ruling out the hypothesis that structural representations fail because they lack mechanistic content. The null also held under scaffold-stratified splits ($\Delta = +0.001$).

An 11.7M-parameter Transformer encoder replicated the null under the same protocol. Morgan fingerprints gave $\Delta = +0.008$, and LINCS and drug-target vectors gave no significant improvement. An omics ablation confirmed that RNA-seq alone captures nearly all predictive signal, with mutations and further modalities adding negligibly (Fig.~\ref{fig:decomp}c). Cell-representation expressiveness is not the bottleneck.

The null replicated across three independent datasets (CTRPv2, BeatAML, and PRISM, Extended Data Table~3). On PRISM, the effect was marginal, consistent with its pharmacologically diverse viability-based assay. The uniformity of the null across all encoding types shifts the question from representation quality to training distribution. Does MoA information fail because it cannot be encoded, or because the training distribution is misaligned?

\subsection*{Training distribution alignment outperforms drug encoding}

If models transfer within but not across MoA classes\cite{herbert2026monotherapy}, is the failure \emph{representational}, reflecting missing MoA knowledge, or \emph{distributional}, reflecting MoA-mismatched training examples? We dissected this by channeling the same MoA information through two routes: as a drug feature (a 24-class one-hot label concatenated with cell features) or as a training distribution constraint ($20\times$ upweighting of same-MoA samples, or within-MoA leave-one-out CV).

\begin{figure}[!htbp]
\centering
\includegraphics[width=\textwidth]{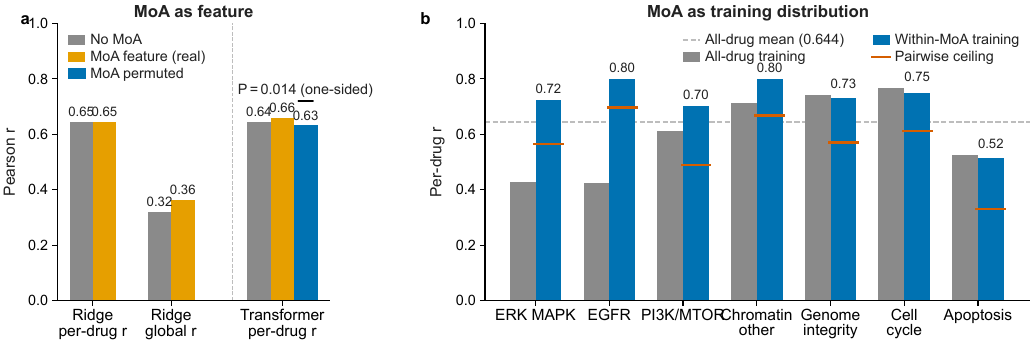}
\caption{\textbf{Training distribution determines the prediction ceiling for targeted kinase inhibitors. Drug encoding does not.}
\textbf{a}, MoA identity as a drug feature. Ridge global $r$ rises by $\MoaOnehotGlobalRDelta$; ridge per-drug $r$ is unchanged. The Transformer's marginal gain with real labels is abolished by label permutation (one-sided Wilcoxon $P = 0.014$).
\textbf{b}, MoA identity as a training distribution. Per-drug $r$ for seven MoA classes under all-drug (gray) vs.\ within-MoA (blue) training. Orange lines, pairwise profile concordance (nearest-neighbor transfer baseline).}
\label{fig:moa}
\end{figure}

Under ridge regression, supplying MoA as a feature left per-drug $r$ unchanged while increasing global $r$ by $\MoaOnehotGlobalRDelta$ (Fig.~\ref{fig:moa}a, Extended Data Table~2). The global $r$ gain is expected because a drug-constant label shifts predicted means without altering relative cell sensitivities. In contrast, $20\times$ upweighting of same-MoA training samples produced large gains for targeted kinase inhibitors, raising ERK MAPK and EGFR per-drug $r$ well above their all-drug baselines (Extended Data Table~2). The same information produces qualitatively different outcomes depending on whether it enters through representation or distribution.

The Transformer partially overcomes this limitation. MoA one-hot gave a marginally significant gain (one-sided Wilcoxon $P = 0.053$) that was abolished by a permuted-label control ($P = 0.014$ for the real-vs.-permuted difference), confirming that the Transformer exploits class identity rather than dimensionality reduction (Fig.~\ref{fig:moa}a). Yet this representational gain remains an order of magnitude smaller than the within-MoA training gains for targeted classes. Distribution alignment is the primary mechanism.

Within-MoA leave-one-out produced the largest gains. ERK MAPK per-drug $r$ rose from $\ErkMapkAllDrug$ to $\ErkMapkMoaR$ across 11 drugs, and EGFR from $\EgfrAllDrug$ to $\EgfrMoaR$ across 7 drugs. Both values exceed the within-class profile concordances, indicating that supervised models extract a generalizable mapping beyond nearest-drug transfer (Fig.~\ref{fig:moa}b).

The improvement was class-specific and graded by pathway specificity. PI3K/MTOR and RTK signaling showed smaller but positive gains, as expected for pharmacologically heterogeneous classes (Supplementary Table~\ref{tab:moa_per_drug_b}). Apoptosis regulation showed no improvement, a genuine biological limit attributable to within-class mechanistic heterogeneity. Cell-division classes were unaffected because all-drug training already captures the pan-cancer fragility signal for these pathways. CTRPv2 replication confirmed the EGFR pattern ($\Delta = \CtrpvTwoEgfrMoaDelta$ for 8 EGFR inhibitors).

\begin{figure}[!htbp]
\centering
\includegraphics[width=\textwidth]{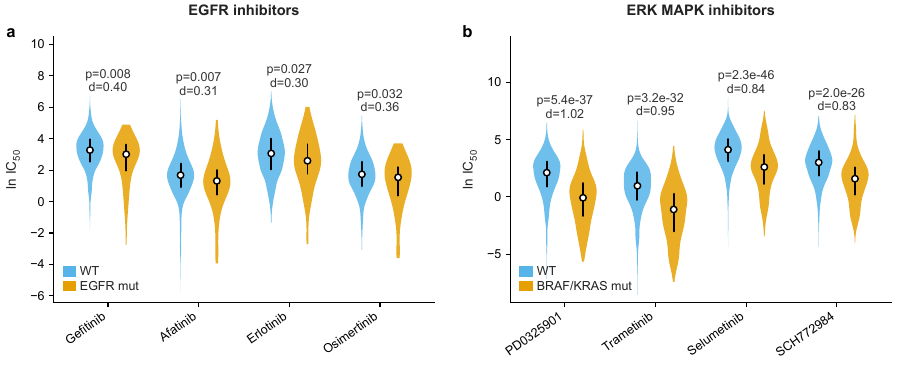}
\caption{\textbf{Oncogenic driver mutations stratify sensitivity to matched inhibitors.}
\textbf{a}, ln IC$_{50}$ distributions for four EGFR inhibitors in \textit{EGFR}-mutant ($n = 60$, orange) vs.\ wild-type (blue) cell lines. Mutant cells are significantly more sensitive (one-sided Mann--Whitney $P < 0.05$, Cohen's $d = 0.30$--$0.40$).
\textbf{b}, ln IC$_{50}$ for four ERK MAPK inhibitors in \textit{BRAF}/\textit{KRAS}-mutant ($n \geq 227$ per drug, orange) vs.\ wild-type (blue) cell lines. Mutant cells are substantially more sensitive (Cohen's $d = 0.83$--$1.02$, $P < 10^{-25}$).}
\label{fig:biomarker}
\end{figure}

Pan-cancer training creates a dominant cell-fragility signal that suppresses the weaker, pathway-specific signals governing targeted kinase inhibitor response. The upweighting result supports this interpretation: gains emerge even when all training drugs are retained, ruling out the alternative that within-MoA training simply presents an easier problem with fewer samples. Consistent with this mechanism, the signals recovered by within-MoA training correspond to known oncogenic drivers\cite{lynch2004activating,davies2002mutations}. \textit{EGFR}-mutant cell lines are significantly more sensitive to all EGFR inhibitors tested (Fig.~\ref{fig:biomarker}a), and \textit{BRAF}/\textit{KRAS}-mutant cell lines show large sensitivity differences to ERK MAPK inhibitors (Cohen's $d > 0.8$, Fig.~\ref{fig:biomarker}b). Pan-cancer co-training dilutes these mutation-stratified signals because most training drugs target different oncogenic pathways.

\subsection*{Response profile matching improves within-drug ranking}

MoA-stratified training requires knowing the drug class in advance. We next ask whether pilot IC$_{50}$ observations of the test drug, which require no class information, can close the gap independently.

\begin{figure}[!htbp]
\centering
\includegraphics[width=\textwidth]{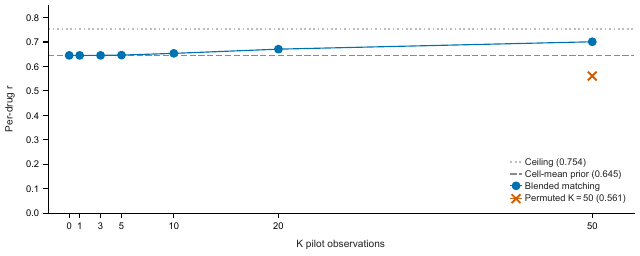}
\caption{\textbf{Pilot profiling consistently improves within-drug cell ranking above $K = 10$ observations.}
Per-drug $r$ vs.\ $K$ pilot observations on GDSC2. The cell-mean prior ($K = 0$, dashed) achieves $r = \CellMeanR$ without any drug-specific information. Blended matching exceeds this at $K \geq 20$ and reaches $r = \KFiftyR$ at $K = 50$, against a replicate-pair ceiling of $\MeasurementCeiling$. A permuted $K = 50$ control (cross) confirms the gain requires correct cell--drug pairing, not response-scale alignment.}
\label{fig:kshot}
\end{figure}

Given $K$ IC$_{50}$ observations for a new drug, response profile matching identifies training drugs with similar response patterns on the observed cells and transfers their sensitivity profiles. At low $K$, too few observations yield unreliable matches and the blended estimator defaults to the cell-mean prior. Above approximately 20 observations, correlation estimates stabilize and matching adds genuine value, reaching per-drug $r = \KFiftyR$ at $K = 50$ (Fig.~\ref{fig:kshot}). A permuted-drug control confirmed that incorrect pairing actively degrades performance, falling below the cell-mean prior.

MoA-stratified training and response matching trade different resources. Within-MoA training requires the drug class label and a same-class training set, whereas response matching requires only pilot observations of the test drug itself. For EGFR, combining the two at $K = 20$ matched within-MoA training alone. For ERK MAPK, response matching alone at $K = 50$ exceeded within-MoA training without requiring class information. Direct observations of the test drug encode its specific response profile more precisely than proxy training on related drugs, and the advantage grows with the number of pilot observations.

\begin{figure}[!htbp]
\centering
\includegraphics[width=\textwidth]{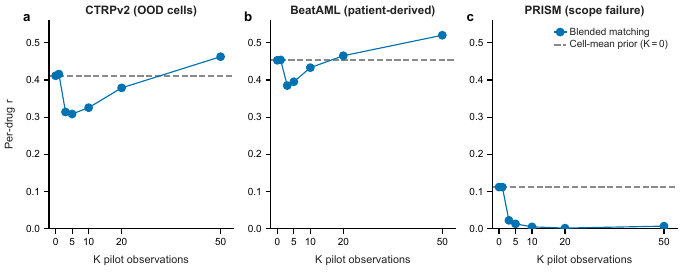}
\caption{\textbf{Response matching generalizes across independent cell lines and patient-derived samples but requires functional drug analogs in the training set.}
\textbf{a}, CTRPv2 (out-of-distribution cells). Per-drug $r$ rises by $\CtrpvTwoKShotLift$ at $K = 50$ using 812 Broad cell lines absent from GDSC2 training.
\textbf{b}, BeatAML (patient-derived). Per-drug $r$ rises from $\BeatAmlKZeroR$ to $\BeatAmlKFiftyR$ at $K = 50$ across 520 AML patients.
\textbf{c}, PRISM (scope failure). The $K$-curve collapses at $K \geq 3$ because PRISM's repurposing library lacks functional analogs among GDSC2's training drugs. Transient dip at $K = 3$--$10$ in panels a and b reflects unreliable estimates from too few observations. Cell-mean prior (dashed line) is the $K = 0$ reference in each panel.}
\label{fig:kcurve}
\end{figure}

Response matching generalizes across independent datasets (Fig.~\ref{fig:kcurve}). On CTRPv2, 812 Broad cell lines absent from GDSC2 training identify informative matches through shared omics features. On patient-derived BeatAML\cite{bottomly2022integrative}, per-drug $r$ rises from $\BeatAmlKZeroR$ to $\BeatAmlKFiftyR$ at $K = 50$, confirming transfer to \textit{ex vivo} patient samples. PRISM's $K$-curve collapses because its pharmacologically diverse repurposing library lacks functional analogs among GDSC2's targeted/cytotoxic training drugs, defining the practical boundary of the approach (Fig.~\ref{fig:kcurve}c).

\section*{Discussion}

The assumption that improved drug representations would raise drug-blind prediction ceilings has driven a decade of encoder development in cancer pharmacogenomics. Our results redefine the constraint. The bottleneck is distributional, not representational. Mechanistically heterogeneous co-training suppresses the pathway-specific signals that any encoder would transmit regardless of its expressiveness, and restructuring the training distribution removes this suppression directly. Independent work has converged on the same null in drug-blind evaluation\cite{herbert2026monotherapy,bernett2025dreval} and in the drug-seen regime\cite{carli2025cellhit}, indicating that encoder development has been addressing the wrong variable.

The mechanism traces to training distribution heterogeneity. Co-training on mechanistically diverse drugs forces the model to learn a response map averaged across MoAs. Pathway-specific cell features that predict sensitivity for EGFR inhibitors conflict with those for topoisomerase inhibitors, and the model resolves this conflict by learning features predictive only of mean potency. A drug encoder, regardless of its expressiveness, feeds into a model whose response map has been smoothed across all MoAs. LINCS L1000 signatures capture a drug's average transcriptional effect on a reference population, not cell-specific response variation. Drug-target vectors are too coarse to resolve the pathway context differences that distinguish sensitive from resistant cell lines. The sole partial exception, MoA categorical identity in the Transformer, was abolished by label permutation, confirming that the residual signal reflects implicit class-conditional partitioning of the training distribution rather than representational content.

Within-MoA training distinguishes artifactual ceilings from genuine biological limits. Aligned training exceeded within-class profile concordances for ERK MAPK and EGFR, indicating that supervised models extract a generalizable mapping beyond profile similarity. Classes with high within-class mechanistic heterogeneity, such as Apoptosis regulation, are not addressable by distribution alignment alone because the training drugs within the class do not share a coherent response pattern to transfer. Response matching provides a complementary route, generalizing to patient-derived data (BeatAML) and to cell lines absent from training (CTRPv2), but requiring functional drug analogs in the reference set.

Checkpoint selection artifacts explain a substantial portion of reported progress as methodological inflation, consistent with the 72\% leakage rate documented across the field\cite{asiaee2026widespread}. Per-drug $r$, the cell-mean prior as a free baseline, and replicate concordance as a ceiling give the field evaluation tools to separate genuine advances from metric optimization.

Both approaches have scope conditions. Response matching requires functional drug analogs in the reference set, and within-MoA training requires mechanistically coherent drug classes with strong pathway dependence. Beyond these scope conditions, all results rest on cell-line IC$_{50}$ values, which abstract away \textit{in vivo} pharmacokinetics, tumor microenvironment effects, clonal heterogeneity, and drug combinations. Whether distribution alignment translates to improved patient response requires prospective validation in clinical cohorts.

The representation--distribution dissociation may extend beyond cancer pharmacogenomics. Any co-training distribution with mechanistic heterogeneity can suppress signals relevant to a specific test condition. Benchmarking should evaluate training distribution alongside representation quality to distinguish improved encoding from improved data curation.

\section*{Methods}

\subsection*{Data}

We used GDSC2\cite{iorio2016landscape} (233 drugs with valid Morgan fingerprints, 687 cancer cell lines with matched RNA and mutation profiles) with RNA-seq gene expression (19,193 protein-coding genes) and somatic mutation profiles (12,301 genes) from the Cancer Cell Line Encyclopedia\cite{ghandi2019next}. The response variable was ln(IC$_{50}$) in $\mu$M. For LINCS experiments, Harmonizome\cite{rouillard2016harmonizome} consensus perturbation profiles (104 GDSC2 drugs matched by name) were PCA-reduced to 64 dimensions. Cross-dataset validation used CTRPv2\cite{seashoreludlow2015} (545 compounds, 812 cell lines, AUC, 66 drugs overlapping GDSC2), BeatAML\cite{bottomly2022integrative} (155 drugs, 520 patients, \textit{ex vivo} AUC, drugs with $\geq 20$ patients), and PRISM Repurposing\cite{corsello2020prism} (1,079 drugs).

\subsection*{Evaluation}

All experiments used $k$-fold drug-blind CV (default $k = 5$, $k = 10$ for PASO splits). For ridge regression, the test fold was held out and the remaining $k-1$ folds were used for training. No validation fold is required. For the Transformer encoder, 10\% of training drugs in each fold were held out as a drug-blind validation set for epoch selection, with the remaining $\sim$90\% used for training. The test fold was evaluated once after model selection. We report global Pearson $r$ (across all test pairs) and per-drug Pearson $r$ (correlation within each test drug across cell lines with $\geq 5$ observations, averaged over drugs). Per-drug $r$ is invariant to additive and multiplicative per-drug transformations of the target.

\subsection*{Models}

The primary model was ridge regression ($\alpha = 1.0$) with RNA PCA-550 + mutation PCA-200 cell features, with drug features appended when specified.

A Transformer encoder\cite{vaswani2017attention} was used for representation ablation experiments. The architecture consists of 4 pre-norm layers (LayerNorm before attention and feedforward), 256-dim hidden size, 8 attention heads, feedforward dimension 1024. Each input modality is projected to 256 dimensions (RNA: Linear($n_{\rm genes} \to 256$), mutations: Linear($n_{\rm mut} \to 256$), drug: Linear($d_{\rm drug} \to 256$)) and summed with a learnable modality-type embedding before self-attention across the three tokens. During training, each omics token is independently zeroed with probability 0.3 (modality dropout). The prediction head mean-pools token outputs, applies LayerNorm, and maps to a scalar (Linear(256, 1)). Total parameters: 11.7M (dominated by the RNA projection). Training used AdamW (lr $= 10^{-3}$, weight decay $10^{-4}$, cosine schedule, $\eta_{\rm min} = 10^{-5}$, with 10-epoch linear warmup), gradient clipping at max norm 1.0, 200 epochs, batch size 256. Drug feature dimensions tested were Morgan FP (2048), LINCS PCA-64, drug-target PCA-256, MoA one-hot (24), and zero-vector (no-drug baseline).

\subsection*{Permuted-MoA control}

To distinguish class-identity information from dimensionality-reduction effects, we constructed a permuted-MoA condition by randomly shuffling drug-to-pathway assignments across all drugs (fixed seed 42) while preserving class-size distribution. The same 10-fold protocol was applied. If the MoA gain reflects class identity, permuted labels should perform at or below the no-drug baseline.

\subsection*{Response profile matching}

Response profile matching extends the collaborative filtering principle\cite{suphavilai2018predicting,zhang2018hybrid} to the drug-blind cold-start setting. Given $K$ observed IC$_{50}$ values for a new drug, we compute Pearson $r$ between the $K$-cell response vector and each training drug's response on the same cells, select the top $N = 5$ training drugs by correlation magnitude, and predict remaining cells as a weighted average of selected training drug profiles, with negative correlations zero-weighted. The blended estimator uses a CV-optimal convex combination of the cell-mean prior and the matched prediction. The blend weight $w$, selected by inner CV per fold, dominates the $K$-dependent sensitivity. At small $K$ the weight collapses to the prior. No model training is required.

\subsection*{MoA-stratified training}

Three conditions. \textit{MoA one-hot}: an $n_{\rm MoA}$-dimensional one-hot vector (GDSC2 Target Pathway, 24 classes) concatenated with cell features in all-drug CV. \textit{MoA-weighted}: all-drug training with same-MoA samples weighted $20\times$ relative to other drugs (weight selected by grid search over $\{1, 2, 5, 10, 20\}$ on validation folds). \textit{Within-MoA CV}: leave-one-drug-out using only same-MoA training drugs, with RNA PCA-550 + mutation PCA-200 features fitted on all training cell lines, ridge ($\alpha = 1.0$), mean and s.d.\ reported over drugs within each class.

\subsection*{PASO reproduction}

PASO's published 10-fold split files were used. PASO-style evaluation retained the checkpoint maximizing test-set Pearson $r$ per epoch (200 epochs, no validation holdout). Fair evaluation held out 10\% of training data as a validation set and selected the checkpoint maximizing validation $r$. The test fold was evaluated once.

\subsection*{Statistical analysis}

Pearson correlations were computed on held-out test folds only. Per-drug $r$ is reported as mean $\pm$ s.d.\ across drugs. Condition differences were assessed by paired Wilcoxon signed-rank test (two-sided unless stated otherwise). $P$ values are exact where sample size permits. Effect sizes are absolute $\Delta r$. Seed sensitivity was assessed over 10 random fold assignments. No multiple-testing correction was applied to exploratory secondary analyses (MoA class comparisons beyond ERK MAPK and EGFR, omics ablation variants). Confirmatory analyses (drug representation null, MoA feature vs.\ distribution dissociation) used pre-specified primary comparisons.

Biomarker associations used binary non-synonymous mutation status from DepMap 24Q4 somatic mutation calls (synonymous variants excluded). Group comparisons (mutant vs.\ wild-type) used a one-sided Mann--Whitney U test with the alternative hypothesis that mutant IC$_{50}$ $<$ wild-type, and pooled Cohen's $d$ as effect size.

\subsection*{Computational protocol}

The headline drug representation ablation and MoA experiments used 10-fold drug-blind CV with 200 epochs (Transformer encoder) or no iterative training (ridge). Exploratory Transformer-encoder ablations (drug feature real/random/zero and PRISM) used reduced configurations (3--5 folds, 15--50 epochs) for computational efficiency. All headline numbers reported in the main text use the full protocol. ChemBERTa\cite{chithrananda2020chemberta} pre-trained SMILES embeddings (768-dim) were evaluated at raw dimensionality and PCA-reduced to 64 and 128 dimensions. For scaffold-stratified evaluation, drugs were partitioned into 5 folds using Bemis--Murcko scaffold clustering\cite{bemis1996properties} (219 unique scaffolds). All experiments were run on an NVIDIA DGX Spark workstation. Ridge regression completed in under one minute per fold, Transformer training in approximately 2 hours per 10-fold run.

\section*{Code availability}

All source code is available at \url{https://github.com/TaekyungHeo/drug-response-distribution}.

\bibliographystyle{unsrtnat}
\bibliography{bibliography}

\clearpage
\section*{Extended Data}
\extdatafigcaptions
\setcounter{figure}{0}

\begin{figure}[!htbp]
\centering
\includegraphics[width=0.95\textwidth]{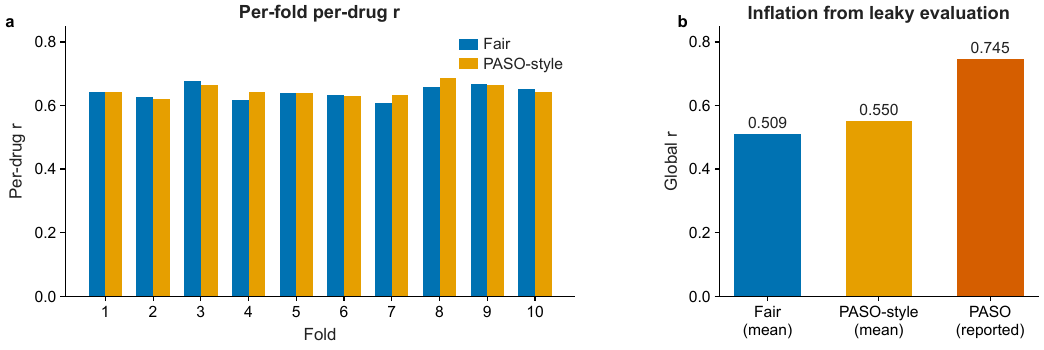}
\caption{\textbf{Test-set checkpoint selection systematically inflates reported drug-blind performance.}
\textbf{a}, Per-fold per-drug $r$ on PASO's 10-fold splits under fair evaluation with a validation holdout vs.\ the published protocol with test-set checkpoint selection. Per-drug $r$ is largely unaffected ($\Delta = +0.004$), because within-drug cell ranking is insensitive to the epoch that maximizes global $r$.
\textbf{b}, Inflation decomposed on global $r$. Test-set snooping adds $\PasoInflation$ to the fair global $r$ mean. Best-fold reporting adds $+0.20$, together reaching \PasoBestFold\ and closely matching the published figure of 0.745.}
\label{fig:paso}
\end{figure}

\clearpage

\begin{table}[!htbp]
\centering
\caption{\textbf{No drug representation improves within-drug cell ranking.} Ridge, 10-fold drug-blind CV, GDSC2. All conditions use $n = 233$ drugs except LINCS ($n = 104$, LINCS-covered subset only). A random 2048-dimensional vector produces the same $\Delta = +0.001$ as the real Morgan fingerprint, confirming that the small positive delta is a mechanical artifact of adding input dimensions under ridge regularization, not drug-specific signal.}
\label{tab:drug_repr_sweep}
\begin{tabular}{lcc}
\toprule
Drug representation & Per-drug $r$ & $\Delta$ vs.\ no drug \\
\midrule
No drug features & $0.645 \pm 0.025$ & --- \\
Morgan fingerprint (ECFP4, 2048-bit) & $0.646 \pm 0.024$ & $+0.001$ \\
Random vector (2048-dim) & $0.646 \pm 0.024$ & $+0.001$ \\
Drug-target profile (binary, 5145 targets) & $0.646 \pm 0.024$ & $+0.001$ \\
LINCS perturbation signature (PCA-64) & $0.665$ & $+0.001$ \\
MoA one-hot (24 classes) & $0.645 \pm 0.025$ & $0.000$ \\
\bottomrule
\end{tabular}
\end{table}

\clearpage

\begin{table}[!htbp]
\centering
\caption{\textbf{MoA as representation vs.\ distribution.} Per-drug $r$ by condition. B vs.\ A tests whether MoA one-hot improves cell ranking ($|\Delta| \leq 0.001$ uniformly). C vs.\ A tests whether MoA-weighted training improves cell ranking (large gains for signaling classes).}
\label{tab:moa_crux}
\resizebox{\textwidth}{!}{%
\begin{tabular}{lcccc}
\toprule
MoA class & A (no drug, uniform) & B (one-hot, uniform) & C (no drug, $20\times$) & $\Delta$(C$-$A) \\
\midrule
ERK MAPK signaling   & 0.427 & 0.428 & 0.673 & $+0.246$ \\
EGFR signaling       & 0.425 & 0.425 & 0.655 & $+0.230$ \\
Apoptosis regulation & 0.524 & 0.524 & 0.547 & $+0.023$ \\
PI3K/MTOR signaling  & 0.610 & 0.610 & 0.703 & $+0.093$ \\
Genome integrity     & 0.742 & 0.742 & 0.747 & $+0.005$ \\
Overall              & 0.645 & 0.645 & 0.687 & $+0.042$ \\
\bottomrule
\end{tabular}%
}
\end{table}

\clearpage

\begin{table}[!htbp]
\centering
\caption{\textbf{Drug-feature null and $K$-shot matching replicate across independent datasets.}
\textbf{a}, Morgan FP vs.\ no drug features, reporting per-drug $r$ and $\Delta$. The null replicates in CTRPv2 and BeatAML. PRISM shows $\Delta = +0.017$, reflecting viability-based cytotoxicity rather than mechanism-specific sensitivity.
\textbf{b}, $K$-shot matching generalizes to CTRPv2 but fails on PRISM due to drug-panel mismatch. BeatAML per-drug $r$ rises from $\BeatAmlKZeroR$ at $K=0$ to $\BeatAmlKFiftyR$ at $K=50$.}
\label{tab:ext_validation}
\begin{subtable}{\textwidth}
\raggedright\textbf{a}\\[4pt]
\centering
\begin{tabular}{lcccc}
\toprule
Dataset & No drug $r$ & Morgan FP $r$ & $\Delta$ & $n_{\text{drugs}}$ \\
\midrule
GDSC2 (primary)  & 0.645 & 0.646 & $+0.001$ & 233 \\
CTRPv2           & 0.412 & 0.417 & $+0.005$ & 545 \\
BeatAML          & 0.453 & 0.454 & $+0.001$ & 155 \\
PRISM            & 0.117 & 0.134 & $+0.017$ & 1079 \\
\bottomrule
\end{tabular}
\end{subtable}

\vspace{1em}

\begin{subtable}{\textwidth}
\raggedright\textbf{b}\\[4pt]
\centering
\begin{tabular}{lcccccc}
\toprule
Dataset & $K=0$ & $K=1$ & $K=3$ & $K=10$ & $K=20$ & $K=50$ \\
\midrule
BeatAML & 0.453 & 0.454 & 0.385 & 0.433 & 0.465 & 0.521 \\
CTRPv2 & 0.411 & 0.416 & 0.314 & 0.325 & 0.379 & 0.463 \\
PRISM  & 0.112 & 0.112 & 0.022 & 0.004 & 0.001 & 0.006 \\
\bottomrule
\end{tabular}
\end{subtable}
\end{table}

\clearpage
\renewcommand{\thetable}{\arabic{table}}
\renewcommand{\thefigure}{\arabic{figure}}
\renewcommand{\theequation}{S\arabic{equation}}
\setcounter{figure}{0}
\setcounter{table}{0}
\setcounter{equation}{0}
\suppfigcaptions

\section*{Supplementary Information}

\begin{figure}[!htbp]
\centering
\includegraphics[width=\textwidth]{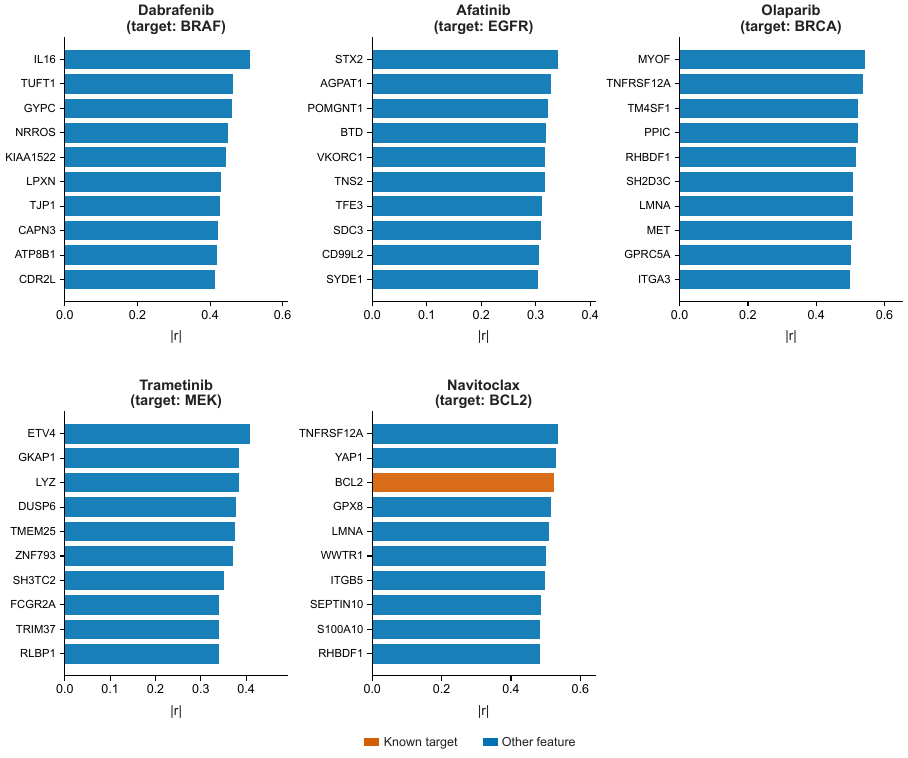}
\caption{\textbf{Top-correlated features reflect lineage rather than chemical structure.} Top 10 most correlated features with IC$_{50}$ for five targeted drugs. Orange bars indicate known molecular targets when present in the top 10. Only Navitoclax places its target (\textit{BCL2}) in the top 10 (3rd). For the remaining four drugs, the known target ranks far below: \textit{EGFR} ranks 29,348th for Afatinib, \textit{BRAF} mutation ranks 9,944th for Dabrafenib. Lineage and cell-state markers dominate instead.}
\label{fig:efig1}
\end{figure}

\clearpage

\subsection*{Supplementary Note 1 | Formal decomposition of global Pearson $r$}

Let $y_{d,c}$ denote the true log IC$_{50}$ for drug $d$ on cell line $c$, and $\hat{y}_{d,c}$ the model prediction.
Decompose each as a sum of a drug-level mean and a within-drug residual:
\begin{equation}
y_{d,c} = \bar{y}_d + \epsilon_{d,c}, \qquad
\hat{y}_{d,c} = \hat{\bar{y}}_d + \hat{\epsilon}_{d,c}
\label{eq:decomp}
\end{equation}
where $\bar{y}_d = \frac{1}{|C_d|}\sum_c y_{d,c}$.

\textbf{Covariance decomposition.}
Expanding $\mathrm{Cov}(y,\hat{y})$:
\begin{equation}
\mathrm{Cov}(y,\hat{y})
= \mathrm{Cov}(\bar{y}_d,\hat{\bar{y}}_d)
+ \underbrace{\mathrm{Cov}(\bar{y}_d,\hat{\epsilon}_{d,c}) + \mathrm{Cov}(\epsilon_{d,c},\hat{\bar{y}}_d)}_{\text{cross-terms}}
+ \mathrm{Cov}(\epsilon_{d,c},\hat{\epsilon}_{d,c})
\end{equation}
When drug-level means and within-drug residuals are approximately uncorrelated in the model's predictions (the assumption holds when the model cannot predict drug-specific deviations), the cross-terms vanish, giving:
\begin{equation}
\mathrm{Cov}(y,\hat{y})
\approx \underbrace{\mathrm{Cov}(\bar{y}_d,\hat{\bar{y}}_d)}_{\alpha:\,\text{between-drug scale}}
+ \underbrace{\mathrm{Cov}(\epsilon_{d,c},\hat{\epsilon}_{d,c})}_{\beta:\,\text{within-drug ranking}}
\label{eq:cov_decomp}
\end{equation}

\textbf{Variance decomposition.}
$\mathrm{Var}(y) = \sigma^2_{\mathrm{between}} + \sigma^2_{\mathrm{within}}$ where $\sigma^2_{\mathrm{between}} = \mathrm{Var}(\bar{y}_d)$ and $\sigma^2_{\mathrm{within}} = \mathrm{Var}(\epsilon_{d,c})$.

\textbf{Global Pearson $r$ as a weighted combination.}
Let $\omega_{\mathrm{b}} = \sigma^{(y)}_{\mathrm{between}} \sigma^{(\hat{y})}_{\mathrm{between}} / (\sigma_y \sigma_{\hat{y}})$ and $\omega_{\mathrm{w}} = \sigma^{(y)}_{\mathrm{within}} \sigma^{(\hat{y})}_{\mathrm{within}} / (\sigma_y \sigma_{\hat{y}})$, with $\omega_{\mathrm{b}} + \omega_{\mathrm{w}} \leq 1$.
Then:
\begin{equation}
r_{\mathrm{global}}
\approx \omega_{\mathrm{b}}\,r_{\mathrm{between}} + \omega_{\mathrm{w}}\,r_{\mathrm{within}}
\label{eq:global_r}
\end{equation}
where $r_{\mathrm{between}} = \mathrm{Cor}(\bar{y}_d,\hat{\bar{y}}_d)$ and $r_{\mathrm{within}} = \mathrm{Cor}(\epsilon_{d,c},\hat{\epsilon}_{d,c})$.

\textbf{Why between-drug variance dominates in GDSC2.}
Drug means $\bar{y}_d$ in GDSC2 span $[-5.2,\, +10.3]$\,ln\,$\mu$M (range 15.5 units), while within-drug standard deviations average $\sim$1.5\,ln\,$\mu$M.
Consequently $\sigma^2_{\mathrm{between}} \gg \sigma^2_{\mathrm{within}}$, so $\omega_{\mathrm{b}} \gg \omega_{\mathrm{w}}$ and global $r$ is dominated by scale prediction accuracy.
A model that predicts only drug means (DummyDrugAvg) achieves $r_{\mathrm{between}} = 1$ and $r_{\mathrm{within}} = 0$, yet obtains high global $r$, the trivial predictor phenomenon\cite{specgame2025}

\textbf{Per-drug $r$ as an estimator of $r_{\mathrm{within}}$.}
The per-drug $r$ defined in the main text (Pearson $r$ within each test drug, averaged across drugs) is a direct estimator of $r_{\mathrm{within}}$: by computing correlation within each drug independently, between-drug mean differences cancel.

\textbf{Empirical consistency.}
Equation~\ref{eq:global_r} predicts that a method improving $r_{\mathrm{between}}$ should increase global $r$ without changing per-drug $r$. A method improving $r_{\mathrm{within}}$ should increase per-drug $r$ with a smaller effect on global $r$, because $\omega_{\mathrm{w}} \ll \omega_{\mathrm{b}}$.
LINCS signatures change global $r$ by $\LincsGlobalRDelta$ on the 104-drug LINCS-covered subset while per-drug $r$ changes by $\LincsPerDrugDelta$ (null), isolating the effect to between-drug scale prediction (here, a degradation rather than an improvement).
Response matching at $K = 50$ increases per-drug $r$ by $\KFiftyLift$ over the cell-mean prior without a corresponding global $r$ increase, consistent with improving $r_{\mathrm{within}}$ only.
These empirical results directly validate the decomposition.

\subsection*{Supplementary Note 2 | Supporting results}

\subsubsection*{Baselines and performance ceilings}

Table~\ref{tab:ceilings} contextualizes the per-drug $r$ achieved by our methods against fundamental baselines and ceilings.

\begin{table}[htbp]
\centering
\caption{\textbf{Per-drug $r$ in context of baselines and measurement ceiling.} Methods are ordered by value. Dagger indicates response matching, which uses $K$ pilot IC$_{50}$ measurements. All other methods are zero-shot.}
\label{tab:ceilings}
\resizebox{\textwidth}{!}{%
\begin{tabular}{lcc}
\toprule
Reference & Per-drug $r$ & Notes \\
\midrule
DummyDrugAvg (drug mean predictor) & 0.000 & Per-drug $r = 0$ by definition \\
Profile concordance (nearest-drug transfer baseline) & 0.528 & Mean across all 22 MoA classes (7 shown in Table~\ref{tab:moa_per_drug_b}) \\
Cell-mean prior (no observations) & 0.645 & 86\% of measurement ceiling \\
Response matching$^\dagger$ ($K = 50$, blended, $N = 5$) & \textbf{0.701} & \textbf{$+0.056$ vs cell-mean, CV blend $w = 0.6$} \\
Estimated within-assay replicate concordance & $\sim$0.754 & Within-protocol noise ceiling \\
\bottomrule
\end{tabular}%
}
\end{table}

Table~\ref{tab:eval_settings} shows ridge regression performance across evaluation settings. The counterintuitive reversal (cell-blind per-drug $r <$ drug-blind) reflects that drug-blind retains all training cell lines, enabling strong cell-state ranking despite not knowing the drug identity.

\begin{table}[htbp]
\centering
\caption{\textbf{Ridge regression performance by evaluation setting.} No drug features. Mixed-set is trivially easy (both drug and cell seen in training) and reported for reference only.}
\label{tab:eval_settings}
\begin{tabular}{lccc}
\toprule
Evaluation setting & Global $r$ & Per-drug $r$ & Note \\
\midrule
Mixed-set (random pair splits) & $\approx$0.89 & --- & Trivial, drug and cell both in training \\
Cell-blind & $\approx$0.22 & 0.438 & \\
Drug-blind & $\approx$0.34 & 0.645 & Primary evaluation setting \\
\bottomrule
\end{tabular}
\end{table}

\subsubsection*{Per-drug performance by pathway}

\begin{figure}[htbp]
\centering
\includegraphics[width=0.75\textwidth]{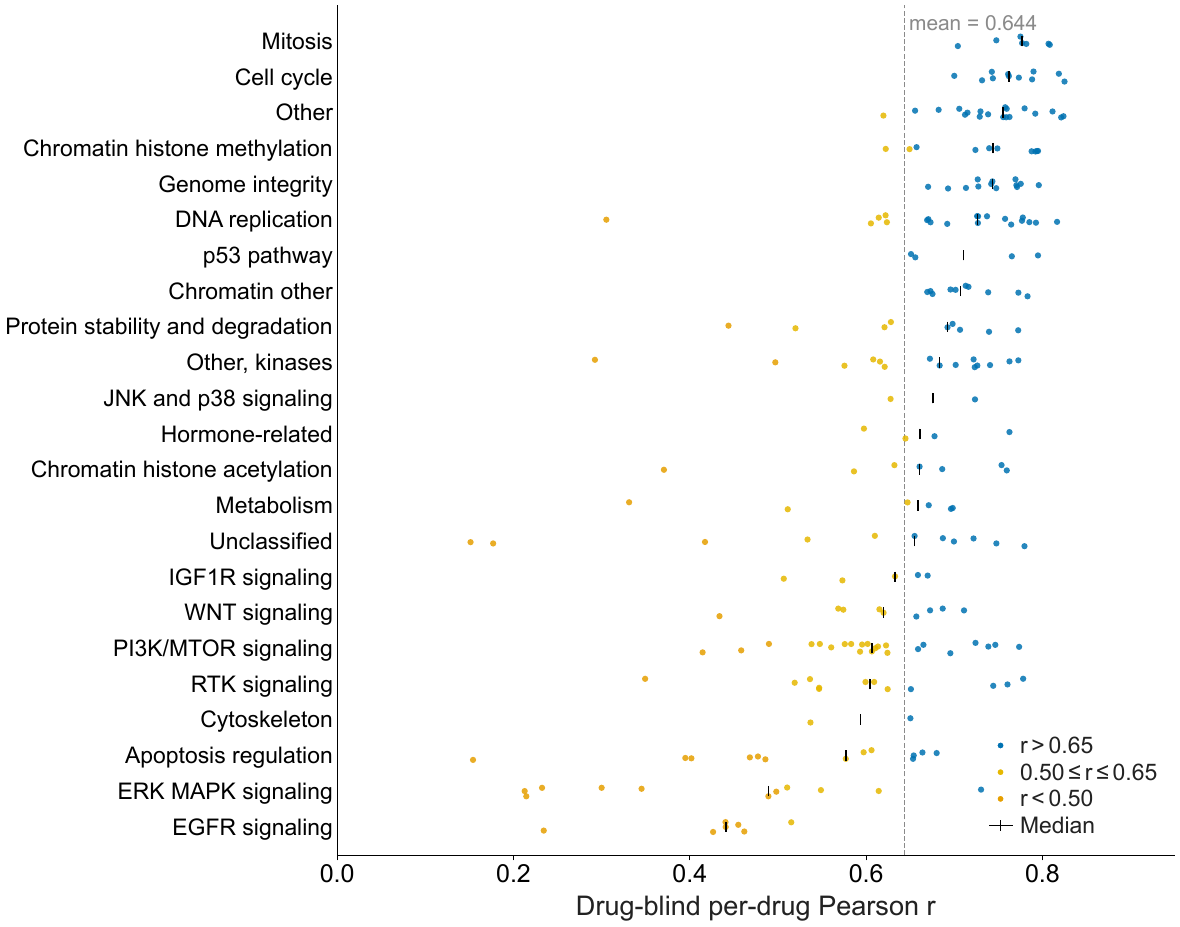}
\caption{\textbf{Per-drug Pearson $r$ by GDSC2 pathway annotation.} Blue, $r > 0.65$. Yellow, $0.50$--$0.65$. Orange, $r < 0.50$. Dashed line, overall mean.}
\label{fig:pathway}
\end{figure}

Drug-blind per-drug $r$ varies substantially across drug pathway classes (Fig.~\ref{fig:pathway}).

\begin{table}[htbp]
\centering
\small
\caption{\textbf{Per-drug Pearson $r$ by GDSC2 Target Pathway under all-drug training.} Ridge, RNA PCA-550 + mutation PCA-200, 10-fold drug-blind CV.}
\label{tab:moa_per_drug_a}
\begin{tabular}{lcr}
\toprule
Target Pathway & Per-drug $r$ (mean $\pm$ s.d.) & $n$ \\
\midrule
Mitosis & $0.772 \pm 0.034$ & 7 \\
Cell cycle & $0.767 \pm 0.036$ & 11 \\
Genome integrity & $0.742 \pm 0.034$ & 13 \\
Chromatin histone methylation & $0.731 \pm 0.063$ & 10 \\
Chromatin other & $0.714 \pm 0.038$ & 10 \\
DNA replication & $0.693 \pm 0.109$ & 20 \\
Other, kinases & $0.648 \pm 0.120$ & 15 \\
WNT signaling & $0.615 \pm 0.079$ & 9 \\
PI3K/MTOR signaling & $0.610 \pm 0.088$ & 23 \\
IGF1R signaling & $0.608 \pm 0.061$ & 5 \\
RTK signaling & $0.605 \pm 0.116$ & 12 \\
Apoptosis regulation & $0.524 \pm 0.143$ & 13 \\
ERK MAPK signaling & $0.427 \pm 0.168$ & 11 \\
EGFR signaling & $0.425 \pm 0.082$ & 7 \\
\midrule
Overall & $0.645$ & 233 \\
\bottomrule
\end{tabular}
\end{table}

The mean per-drug $r$ delta (Morgan FP $-$ no drug) of $+0.001$ is uniformly near zero across all 233 drugs (97\% show $|\Delta| \leq 0.01$, range $[-0.014, +0.018]$). Grouped by pathway class, even the largest pathway mean ($+0.006$ for RTK signaling) is $9\times$ smaller than the $K$-shot gain at $K = 50$.

\subsubsection*{Within-MoA training results and profile concordance}

\begin{table}[htbp]
\centering
\small
\caption{\textbf{Within-class profile concordance and within-MoA training results.} Profile concordance is estimated from pairwise per-drug profile correlations within each MoA class. Within-MoA $r$ uses leave-one-drug-out CV restricted to same-MoA drugs.}
\label{tab:moa_per_drug_b}
\begin{tabular}{lcccc}
\toprule
MoA class & $n$ & All-drug $r$ & Profile concordance & Within-MoA $r$ \\
\midrule
ERK MAPK signaling & 11 & 0.427 & $\ErkMapkProfileConcordance \pm \ErkMapkProfileConcordanceStd$ & $0.723$ \\
EGFR signaling & 7 & 0.425 & $0.697 \pm 0.051$ & $0.799$ \\
PI3K/MTOR signaling & 23 & 0.610 & $0.489 \pm 0.147$ & $0.702$ \\
RTK signaling & 12 & 0.605 & $0.454 \pm 0.142$ & $0.640$ \\
Apoptosis regulation & 13 & 0.524 & $0.331 \pm 0.164$ & $0.515$ ($\Delta = -0.009$) \\
Genome integrity & 13 & 0.742 & $0.570 \pm 0.083$ & $0.730$ \\
Mitosis & 7 & 0.772 & $0.713 \pm 0.090$ & $0.788$ \\
\bottomrule
\end{tabular}
\end{table}

For ERK MAPK and EGFR, within-MoA training recovers per-drug $r$ that substantially exceeds the nearest-neighbor transfer baseline. Apoptosis regulation is unchanged ($\Delta = -0.009$), reflecting a genuine biological limit.

\subsection*{Supplementary Note 3 | Methods and robustness}

\subsubsection*{$K$-shot blend calibration}

\begin{table}[htbp]
\centering
\caption{\textbf{CV-selected blend weight and per-drug $r$ by $K$.} $w = 0$ corresponds to cell-mean prior only and $w = 1$ to matching only. Permuted-drug control at $K = 50$ yields $r = 0.561$.}
\label{tab:kshot_curve}
\begin{tabular}{cccc}
\toprule
$K$ & CV blend weight ($w$) & Per-drug $r$ & $\Delta$ vs.\ prior \\
\midrule
0  & ---  & 0.645 & --- \\
1  & 0.1  & 0.645 & $0.000$ \\
3  & 0.0  & 0.645 & $0.000$ \\
5  & 0.1  & 0.646 & $+0.001$ \\
10 & 0.2  & 0.653 & $+0.008$ \\
20 & 0.3  & 0.670 & $+0.025$ \\
50 & 0.5--0.6 & \textbf{0.701} & $\mathbf{\KFiftyLift}$ \\
\bottomrule
\end{tabular}
\end{table}

\subsubsection*{Cell representation alternatives}

\begin{table}[htbp]
\centering
\caption{\textbf{Cell representation alternatives all yield per-drug $r \approx 0.645$.} Ridge, no drug features, 10-fold drug-blind PASO CV on GDSC2.}
\label{tab:cell_repr}
\begin{tabular}{lcc}
\toprule
Cell representation & Dimensions & Per-drug $r$ \\
\midrule
RNA PCA(550) + mutation PCA(200) [baseline] & 750 & 0.645 \\
RNA PCA(1000) + mutation PCA(200) & 1,200 & 0.645 \\
KEGG pathway scores & 1,284 & 0.645 \\
RNA PCA(550) + pathway features & 1,834 & 0.645 \\
RNA PCA(100) & 100 & 0.645 \\
Pathway PCA(200) & 200 & 0.645 \\
Mutations PCA(200) only & 200 & 0.645 \\
\bottomrule
\end{tabular}
\end{table}

\end{document}